\documentclass[runningheads]{llncs}
\usepackage{graphicx}
\usepackage{cite}
\usepackage{amsmath,amssymb,amsfonts}
\usepackage{algorithmic}
\usepackage{graphicx}
\usepackage{textcomp}
\usepackage{multirow}
\usepackage{subcaption}
\usepackage{listings}
\usepackage{hyperref}
\usepackage{dirtree}
\usepackage[table,xcdraw]{xcolor}

\definecolor{codegreen}{rgb}{0,0.6,0}
\definecolor{codegray}{rgb}{0.5,0.5,0.5}
\definecolor{codepurple}{rgb}{0.58,0,0.82}
\definecolor{backcolour}{rgb}{0.95,0.95,0.92}

\lstdefinestyle{mystyle}{
    backgroundcolor=\color{backcolour},   
    commentstyle=\color{codegreen},
    keywordstyle=\color{magenta},
    numberstyle=\tiny\color{codegray},
    stringstyle=\color{codepurple},
    basicstyle=\ttfamily\footnotesize,
    breakatwhitespace=false,         
    breaklines=true,                 
    captionpos=b,                    
    keepspaces=true,                 
    numbers=left,                    
    numbersep=5pt,                  
    showspaces=false,                
    showstringspaces=false,
    showtabs=false,                  
    tabsize=2
}

\def\BibTeX{{\rm B\kern-.05em{\sc i\kern-.025em b}\kern-.08em
    T\kern-.1667em\lower.7ex\hbox{E}\kern-.125emX}}

\newcommand{\ttilde}{{\raise.17ex\hbox{$\scriptstyle\sim$}} }

\renewcommand{\arraystretch}{1.05}

\begin{document}

\title{Pre- and Post-Treatment Glioma Segmentation with the Medical Imaging Segmentation Toolkit}
\titlerunning{Pre and Post-Treatment Glioma Segmentation with MIST}
\author{
Adrian Celaya\inst{1,2} \and
Tucker Netherton\inst{1} \and
Dawid Schellingerhout\inst{1} \and
Caroline Chung\inst{1} \and
Beatrice Riviere\inst{2} \and
David Fuentes\inst{1}
}
\authorrunning{A. Celaya et al.}

\institute{
The University of Texas MD Anderson Cancer Center, Houston TX 77030 \and
Rice University, Houston TX 77005
}

\maketitle 
\begin{abstract}
Medical image segmentation continues to advance rapidly, yet rigorous comparison between methods remains challenging due to a lack of standardized and customizable tooling. In this work, we present the current state of the Medical Imaging Segmentation Toolkit (MIST), with a particular focus on its flexible and modular postprocessing framework designed for the BraTS 2025 pre- and post-treatment glioma segmentation challenge. Since its debut in the 2024 BraTS adult glioma post-treatment segmentation challenge, MIST’s postprocessing module has been significantly extended to support a wide range of transforms, including removal or replacement of small objects, extraction of the largest connected components, and morphological operations such as hole filling and closing. These transforms can be composed into user-defined strategies, enabling fine-grained control over the final segmentation output. We evaluate three such strategies - ranging from simple small-object removal to more complex, class-specific pipelines - and rank their performance using the BraTS ranking protocol. Our results highlight how MIST facilitates rapid experimentation and targeted refinement, ultimately producing high-quality segmentations for the BraTS 2025 challenge. MIST remains open source and extensible, supporting reproducible and scalable research in medical image segmentation.
\end{abstract}

\section{Introduction}
Medical imaging segmentation is a highly active area of research. Since its introduction in 2015, the U-Net architecture has received nearly 90,000 citations on Google Scholar \cite{unet, unet-3d}. Other architectures and frameworks like nnUNet have achieved state-of-the-art accuracy in several medical imaging benchmarks such as the Brain Tumor Segmentation (BraTS) and Medical Segmentation Decathlon (MSD) challenges \cite{isensee2021nnu, isensee2021nnubrats, antonelli2022medical}. Since the introduction of nnUNet in 2018, a variety of innovative segmentation methods have emerged, including transformer-based architectures \cite{vaswani2017attention} and advanced loss functions such as boundary and generalized surface losses \cite{bl, celaya2023generalized}. Despite these advancements, there remains a lack of standardized tools that support consistent, customizable, and reproducible comparisons across segmentation methods. Discrepancies in reported performance - where several works claim improvements over nnUNet \cite{tang2022self, chen2021transunet, hatamizadeh2021swin, zhou2021nnformer, cao2022swin} while others challenge these claims \cite{isensee2024nnurevisit} - highlight the need for a common experimental framework.

To address this need, we introduced the Medical Imaging Segmentation Toolkit (MIST), a modular, end-to-end framework for training, testing, and evaluating deep learning-based segmentation methods in a standardized and reproducible manner. Using MIST, we achieved third place in the 2024 BraTS adult glioma post-treatment challenge, validating the framework's effectiveness in competitive, real-world benchmarks.

Since that submission, we have overhauled several key components of MIST to further support rapid experimentation and evaluation. These updates include a redesigned model interface supporting more flexible architecture and loss function customization, an enhanced inference engine with support for different ensembling and test-time augmentation strategies, and an entirely restructured postprocessing module that enables class-specific strategies such as morphological filtering and label reassignment. In this work, we focus on MIST’s redesigned postprocessing framework and its role in producing high-quality segmentations for the BraTS 2025 pre- and post-treatment glioma segmentation challenge \cite{karargyris2023federated, baid2021rsna, menze2014multimodal, bakas2017advancing, bakas2017segmentation, bakas2017segmentation2, de20242024}. We present a set of postprocessing strategies composed of modular transforms, such as small object removal, hole filling, and class-specific component filtering. We also evaluate their impact on segmentation performance using the BraTS ranking protocol.

Our results show that well-designed postprocessing strategies can improve average segmentation metrics and help address specific failure modes. By enabling fine-grained control over the final segmentation output, MIST allows researchers to iterate quickly and target improvements for specific classes or failure modes. The framework remains open-source and extensible, providing a robust foundation for reproducible and scalable research in 3D medical image segmentation.

\section{Methods} \label{sec:methods}
In this section, we begin by detailing our training protocols used to create our baseline models and results. We then describe MIST's postprocessing module, define our selected postprocessing strategies, and explain the BraTS ranking system.

\subsection{BraTS 2025 Training Protocols} \label{sec:methods-brats}
Our choice of architecture is the Pocket nnUNet with deep supervision (two supervision heads) and residual convolution blocks \cite{celaya2022pocketnet, isensee2021nnu}. MIST automatically selects a patch size of 128$\times$128$\times$128. We use two NVIDIA H100 GPUs with a batch size of four uniformly distributed across the GPUs. Additionally, we use L2 regularization with a penalty parameter equal to $1\times 10^{-5}$. Our choice of loss function is the Dice with Cross Entropy loss. We train each model for 10,000 epochs per fold. We use a cosine learning rate schedule with an initial learning rate of 0.001 \cite{loshchilov2016sgdr}. Within each fold, we set aside 2.5\% of the training data as a validation set to select the best model. Once training is complete, MIST runs inference on the challenge validation data using test time augmentation (flipping along each axis and averaging each prediction) for each model. Inference uses sliding windows with an overlap of 0.5 with Gaussian blending ($\sigma=0.125$). The predictions from all five models are averaged to produce a final prediction. All other options and hyperparameters are left at their default values. Please refer to MIST's \href{https://mist-medical.readthedocs.io/en/latest/}{documentation} for these values.

\subsection{Postprocessing Strategies} \label{sec:postprocessing}
To facilitate reproducible and flexible mask refinement, we redesigned the MIST postprocessing module around a strategy-based architecture that enables users to compose class-specific transformation pipelines in a modular and declarative manner. Each strategy is defined via a JSON configuration file that specifies a sequence of transformations to apply to predicted segmentation masks. Available transformations include the removal of small objects, replacement of small components with a specified label, extraction of the top-$k$ largest connected components, and morphological operations such as hole filling and closing. These transforms can be configured to act globally or selectively on specific segmentation labels (i.e., tumor subregions), and applied either jointly or sequentially across classes.

The user can control parameters such as size thresholds, replacement labels, and morphological operation settings directly through the strategy file, allowing for rapid experimentation without modifying code. This design separates postprocessing logic from the training pipeline, enabling efficient tuning of segmentation quality in the final prediction phase. Furthermore, the system is fully extensible—custom postprocessing transforms can be registered via a decorator-based interface within MIST’s internal transform registry. Collectively, this module provides a lightweight and highly configurable framework for optimizing segmentation outputs, which is particularly beneficial in settings like BraTS, where postprocessing can have a substantial impact on both clinical relevance and leaderboard ranking.

In this submission, we compare three different postprocessing strategies:
\begin{itemize}
    \item \textit{Strategy 1}: Remove small objects in the RC class. We set the size threshold to 100 voxels.
    \item \textit{Strategy 2}: Apply Strategy 1, retain the largest connected component in the RC class, and fill holes in the WT class with the SNFH class.
    \item \textit{Strategy 3}: Replace small objects in the ET and RC classes with the SNFH class and then remove small objects from the SNFH class. The size thresholds for the RC, ET, and SNFH classes are 100, 100, and 64 voxels, respectively.
\end{itemize}

\subsection{Strategy Ranking} \label{sec:ranking}
We evaluate our baseline results alongside our three postprocessing strategies using the BraTS ranking system, which is designed to fairly assess segmentation methods across a set of test patients using multiple metrics. Specifically, we calculate the global Dice score and the 95th percentile Hausdorff distance (HD95) for the non-enhancing tumor core (NETC - label 1), surrounding non-enhancing FLAIR hyperintensity (SNFH - label 2), enhancing tissue (ET - label 3), resection cavity (RC - label 4), tumor core (TC - labels 1 and 3), and whole tumor (WT - labels 1, 2, and 3) classes for each prediction resulting from our five-fold cross-validation. For each patient, metric, and segmentation class, each strategy is ranked based on its segmentation performance. For example, for the first patient using the Dice coefficient on the NETC class, Strategy 3 might receive a rank of 1, while the baseline might receive a rank of 2, and so forth.

This ranking process is repeated across all patients, metrics, and classes, resulting in individual ranks for each strategy. We then calculate a per-patient average rank, which summarizes the overall performance of each strategy for that patient. Finally, we obtain a global rank by averaging the per-patient ranks across the entire dataset. The strategy with the lowest global average rank is considered the top performer. This rank-based evaluation emphasizes consistency and robustness across both patients and evaluation metrics, rather than relying solely on aggregated performance values (i.e., the mean of each metric).

\section{Results} \label{sec:results}
\begin{figure}[ht!]
    \centering
    \includegraphics[width=0.95\textwidth]{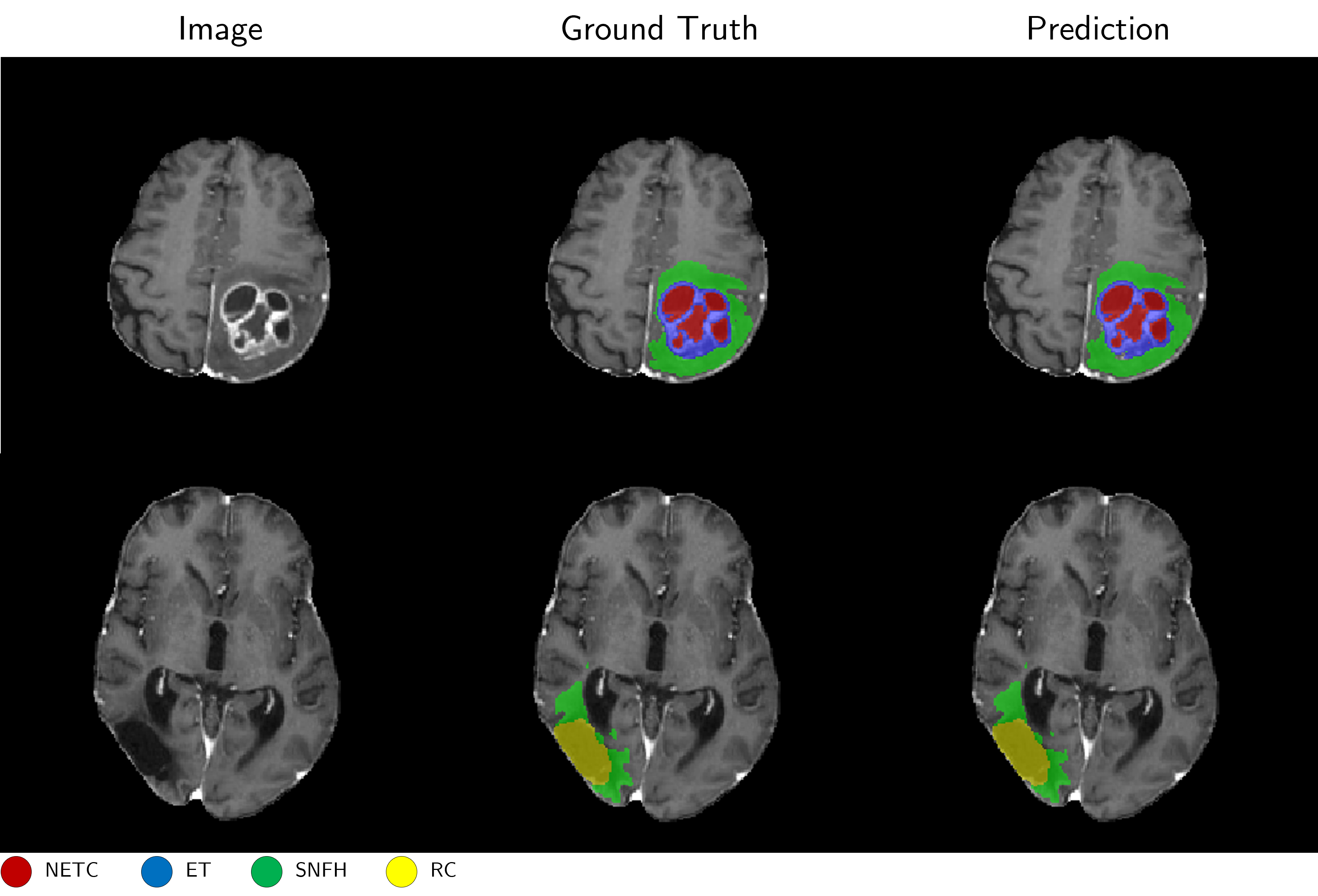}
    \caption{From left to right, a slice of the contrast enhanced T1-weighted image, the ground truth  overlaid on the image, and the baseline prediction overlaid the image for a pre-treatment case (top) and a post-treatment case (bottom).\label{fig:example-predictions}}
\end{figure}

We use MIST and the training parameters described in Section~\ref{sec:methods-brats} to perform a five-fold cross-validation with the BraTS Glioma Segmentation on Pre- and Post-treatment MRI challenge training dataset. Figure~\ref{fig:example-predictions} shows an example of a baseline (i.e., no postprocessing) prediction from a pre-treatment (top) and post-treatment (bottom) case. The three postprocessing strategies described in Section~\ref{sec:postprocessing} are applied to the predictions from the five-fold cross-validation. The accuracy of the baseline and postprocessed predictions is summarized in Table~\ref{tab:results}. Here, we see that Strategy 3 has the best average accuracy with respect to the Dice score for the SNFH, ET, TC, and WT classes, while tying Strategy 1 and the baseline for the best RC and NETC Dice scores, respectively. With respect to the Hausdorff distance, Strategy 3 achieves the best accuracy (i.e., lowest values) for the ET and TC classes, while tying the baseline and Strategy 1 for the lowest NETC and RC distances, respectively. However, Strategy 3 appears to achieve a worse average Hausdorff distance for the SNFH and WT classes compared to the baseline and Strategy 1. Strategy 2 appears to achieve the second-best performance for the average Dice and Hausdorff distance for the RC class while generally matching the baseline and Strategy 1 for the other classes. 

We next apply the BraTS ranking system to our baseline and postprocessed results. Table~\ref{tab:avg-rank-cv} shows the average global rank across all classes and metrics. Despite Strategy 3's meaningful improvements compared to the baseline and other strategies with respect to the average metrics, it does not show an improvement when ranked against the other strategies using the BraTS ranking system. Indeed, when we compare the baseline and our postprocessing strategies using this ranking system, we see that Strategy 3 has the worst (i.e., highest) average rank.

Table~\ref{tab:results-validation} shows the mean and standard deviation of the global Dice, global HD95, lesion-wise (LW) Dice, and LW HD95 for the predictions on the validation set from the baseline model and postprocessed predictions from Strategies 1 and 3. Table~\ref{tab:avg-rank-val} shows the average global rank of these different strategies across all classes and LW metrics for the validation set. Like with the cross-validation results, we see that Strategy 3 generally improves the overall averages of the metrics, but that this improvement in the raw averages of each metric does not translate to a high ranking under the BraTS ranking system. With the validation set, we see that the baseline predictions (i.e., no postprocessing) achieve the highest average global ranking.


\begin{table}[ht!]
\centering
\resizebox{0.9\textwidth}{!}{%
\setlength{\tabcolsep}{20pt} 
\renewcommand{\arraystretch}{1.05} 
\begin{tabular}{llll}
\hline
Strategy & Class & Dice & HD95 (mm) \\ \hline
\rowcolor[HTML]{EFEFEF} 
\cellcolor[HTML]{EFEFEF} & NETC & \cellcolor{yellow}0.7918 (0.3058) & \cellcolor{yellow}22.007 (75.901) \\
\rowcolor[HTML]{EFEFEF} 
\cellcolor[HTML]{EFEFEF} & SNFH & 0.8686 (0.1380) & \cellcolor{yellow}6.8846 (25.077) \\
\rowcolor[HTML]{EFEFEF} 
\cellcolor[HTML]{EFEFEF} & ET & 0.8049 (0.2755) & 26.069 (83.223) \\
\rowcolor[HTML]{EFEFEF} 
\cellcolor[HTML]{EFEFEF} & RC & 0.8234 (0.3080) & 31.475 (93.119) \\
\rowcolor[HTML]{EFEFEF} 
\cellcolor[HTML]{EFEFEF} & TC & 0.8204 (0.2735) & 24.252 (78.966) \\
\rowcolor[HTML]{EFEFEF} 
\multirow{-6}{*}{\cellcolor[HTML]{EFEFEF}Baseline} & WT & 0.9067 (0.1134) & \cellcolor{yellow}6.5513 (22.732) \\ \hline

 & NETC & 0.7918 (0.3058) & 22.007 (75.901) \\
 & SNFH & 0.8686 (0.1380) & 6.8846 (25.077) \\
 & ET & 0.8049 (0.2755) & 26.069 (83.223) \\
 & RC & \cellcolor{yellow}0.8441 (0.2867) & \cellcolor{yellow}26.782 (86.005) \\
 & TC & 0.8204 (0.2735) & 24.252 (78.966) \\
\multirow{-6}{*}{Strategy 1} & WT & 0.9067 (0.1134) & 6.5513 (22.732) \\ \hline

\rowcolor[HTML]{EFEFEF} 
\cellcolor[HTML]{EFEFEF} & NETC & 0.7918 (0.3058) & 22.007 (75.901) \\
\rowcolor[HTML]{EFEFEF} 
\cellcolor[HTML]{EFEFEF} & SNFH & 0.8686 (0.1380) & 6.8846 (25.077) \\
\rowcolor[HTML]{EFEFEF} 
\cellcolor[HTML]{EFEFEF} & ET & 0.8049 (0.2755) & 26.069 (83.223) \\
\rowcolor[HTML]{EFEFEF} 
\cellcolor[HTML]{EFEFEF} & RC & 0.8401 (0.2941) & 28.780 (88.832) \\
\rowcolor[HTML]{EFEFEF} 
\cellcolor[HTML]{EFEFEF} & TC & 0.8204 (0.2735) & 24.252 (78.966) \\
\rowcolor[HTML]{EFEFEF} 
\multirow{-6}{*}{\cellcolor[HTML]{EFEFEF}Strategy 2} & WT & 0.9067 (0.1134) & 6.5513 (22.732) \\ \hline

 & NETC & 0.7918 (0.3058) & 22.007 (75.901) \\
 & SNFH & \cellcolor{yellow}0.8688 (0.1373) & 7.0278 (24.403) \\
 & ET & \cellcolor{yellow}0.8229 (0.2596) & \cellcolor{yellow}23.625 (78.843) \\
 & RC & 0.8441 (0.2867) & 26.782 (86.005) \\
 & TC & \cellcolor{yellow}0.8348 (0.2606) & \cellcolor{yellow}22.528 (75.647) \\
\multirow{-6}{*}{Strategy 3} & WT & \cellcolor{yellow}0.9071 (0.1119) & 6.5872 (21.991) \\ \hline
\end{tabular}%
}
\caption{Mean and standard deviation of the Dice and HD95 for each strategy and segmentation class. The best Dice (highest) and HD95 (lowest) per class are highlighted in \cellcolor{yellow}yellow. Ties are resolved in favor of the earlier-listed strategies.\label{tab:results}}
\end{table}


\begin{table}[ht!]
\centering
\resizebox{\textwidth}{!}{%
\setlength{\tabcolsep}{7.5pt}
\renewcommand{\arraystretch}{1.15}
\begin{tabular}{llllll}
\hline
Stg. & Class & Dice & HD95 & LW Dice & LW HD95 \\ \hline
\rowcolor{yellow} 
\cellcolor[HTML]{EFEFEF} & \cellcolor[HTML]{EFEFEF}NETC & 0.7114 (0.3556) & 42.902 (105.52) & 0.7450 (0.3335) & 37.369 (96.851) \\
\rowcolor[HTML]{EFEFEF} 
\cellcolor[HTML]{EFEFEF} & SNFH & 0.8608 (0.1543) & \cellcolor{yellow}7.1704 (25.163) & 0.8088 (0.2064) & \cellcolor{yellow}24.313 (57.523) \\
\rowcolor[HTML]{EFEFEF} 
\cellcolor[HTML]{EFEFEF} & ET & 0.7807 (0.2847) & \cellcolor{yellow}23.670 (77.327) & 0.7641 (0.2871) & 35.802 (89.441) \\
\rowcolor[HTML]{EFEFEF} 
\cellcolor[HTML]{EFEFEF} & RC & 0.8445 (0.3058) & 25.010 (82.263) & 0.8494 (0.2993) & 29.059 (88.479) \\
\rowcolor[HTML]{EFEFEF} 
\cellcolor[HTML]{EFEFEF} & TC & 0.7925 (0.2855) & 24.582 (77.193) & 0.7664 (0.2938) & 38.956 (89.077) \\
\rowcolor[HTML]{EFEFEF} 
\multirow{-6}{*}{\cellcolor[HTML]{EFEFEF}Base} & WT & 0.9219 (0.0867) & 5.9289 (18.755) & 0.8663 (0.1754) & 23.190 (55.493) \\ \hline
 & NETC & 0.7114 (0.3556) & 42.902 (105.52) & 0.7450 (0.3335) & 37.369 (96.851) \\
 & SNFH & 0.8608 (0.1543) & 7.1704 (25.163) & 0.8088 (0.2064) & 24.313 (57.523) \\
 & ET & 0.7807 (0.2847) & 23.670 (77.327) & 0.7641 (0.2871) & 35.802 (89.441) \\
 & RC & \cellcolor{yellow}0.8484 (0.3026) & \cellcolor{yellow}23.350 (79.303) & \cellcolor{yellow}0.8533 (0.2959) & \cellcolor{yellow}27.804 (86.984) \\
 & TC & 0.7925 (0.2855) & 24.582 (77.193) & 0.7664 (0.2938) & 38.956 (89.077) \\
\multirow{-6}{*}{Stg. 1} & WT & 0.9219 (0.0867) & 5.9289 (18.755) & 0.8663 (0.1754) & 23.190 (55.493) \\ \hline
\rowcolor[HTML]{EFEFEF} 
\cellcolor[HTML]{EFEFEF} & \cellcolor[HTML]{EFEFEF}NETC & 0.7114 (0.3556) & 42.902 (105.52) & 0.7450 (0.3335) & 37.369 (96.851) \\
\rowcolor[HTML]{EFEFEF} 
\cellcolor[HTML]{EFEFEF} & SNFH & \cellcolor{yellow}0.8612 (0.1556) & 7.3227 (25.249) & \cellcolor{yellow}0.8094 (0.2069) & 24.533 (57.427) \\
\rowcolor[HTML]{EFEFEF} 
\cellcolor[HTML]{EFEFEF} & ET & \cellcolor{yellow}0.7892 (0.2789) & 23.760 (77.282) & \cellcolor{yellow}0.7714 (0.2843) & \cellcolor{yellow}35.365 (89.580) \\
\rowcolor[HTML]{EFEFEF} 
\cellcolor[HTML]{EFEFEF} & RC & 0.8484 (0.3026) & 23.350 (79.303) & 0.8533 (0.2959) & 27.804 (86.984) \\
\rowcolor{yellow} 
\cellcolor[HTML]{EFEFEF} & \cellcolor[HTML]{EFEFEF}TC & 0.8012 (0.2739) & 23.066 (73.936) & 0.7723 (0.2869) & 37.949 (87.151) \\
\rowcolor{yellow} 
\multirow{-6}{*}{\cellcolor[HTML]{EFEFEF} Stg. 3} & \cellcolor[HTML]{EFEFEF}WT & 0.9226 (0.0858) & 5.8714 (18.738) & 0.8701 (0.1718) & 21.994 (53.930) \\ \hline
\end{tabular}%
}
\caption{Mean and standard deviation of the Dice, HD95, lesion-wise (LW) Dice, and LW HD95 for the baseline, Strategy 1, and Strategy 3 predictions on the validation set. The best values for each metric are highlighted in yellow. Ties are resolved in favor of the earlier-listed strategies.\label{tab:results-validation}}
\end{table}

\begin{table}[ht!]
\centering
\renewcommand{\arraystretch}{1.05}
\setlength{\tabcolsep}{8pt}
\begin{subtable}[t]{0.48\textwidth}
\centering
\begin{tabular}{ll}
\hline
Strategy & Average Rank \\ \hline
\rowcolor[HTML]{EFEFEF} Strategy 1 & \cellcolor{yellow}2.470288 \\
\rowcolor[HTML]{FFFFFF} Strategy 2 & 2.474945 \\
\rowcolor[HTML]{EFEFEF} Base       & 2.491658 \\
\rowcolor[HTML]{FFFFFF} Strategy 3 & 2.563109 \\ \hline
\end{tabular}
\caption{Cross-validation.}
\label{tab:avg-rank-cv}
\end{subtable}%
\hfill
\begin{subtable}[t]{0.48\textwidth}
\centering
\begin{tabular}{ll}
\hline
Strategy & Average Rank \\ \hline
\rowcolor[HTML]{EFEFEF} Base       & \cellcolor{yellow}1.991810 \\
\rowcolor[HTML]{FFFFFF} Strategy 1 & 1.995547 \\
\rowcolor[HTML]{EFEFEF} Strategy 3 & 2.012643 \\ \hline
\end{tabular}
\caption{Validation set.}
\label{tab:avg-rank-val}
\end{subtable}
\caption{Average global rank for each strategy. (a) Cross-validation set. (b) Validation set. Lower values indicate better performance.}
\label{tab:avg-rank-combined}
\end{table}

\begin{figure}[ht!]
\centering
\begin{minipage}[c]{0.55\textwidth}
    \includegraphics[width=\linewidth]{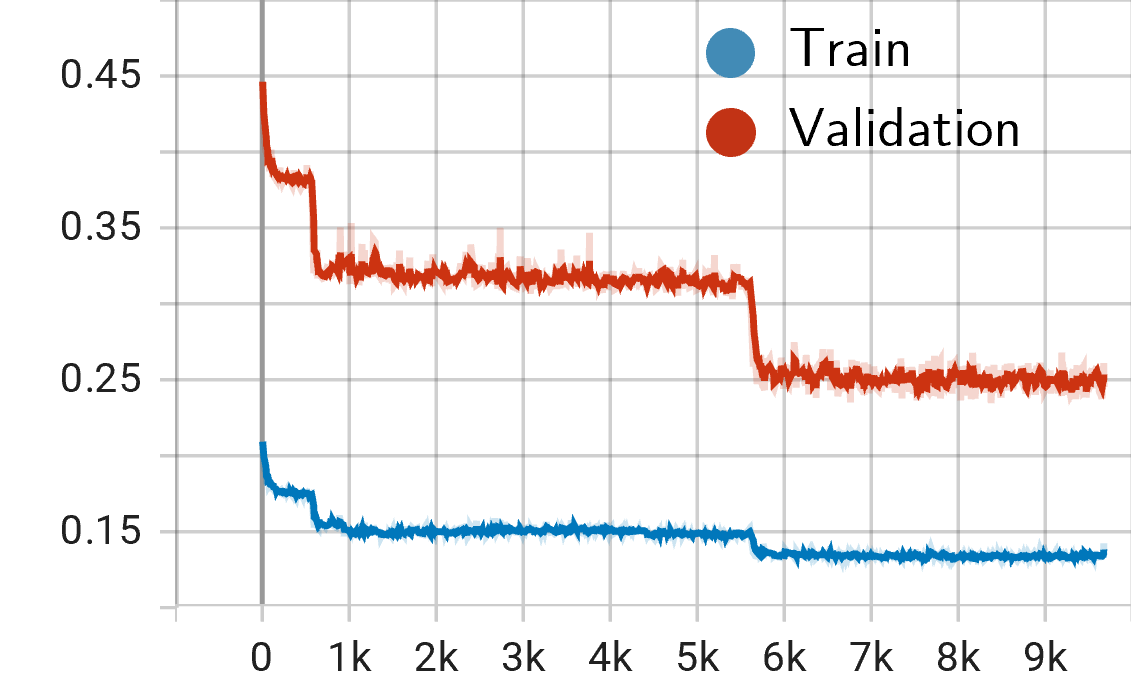}
\end{minipage}%
\hfill
\begin{minipage}[c]{0.40\textwidth}
    \captionof{figure}{Example of train (blue) and validation (red) loss curves for our model. The training loss is the mean of the Dice and cross entropy loss functions and the validation loss is the Dice loss function without the background class.\label{fig:loss-curves}}
\end{minipage}
\end{figure}

\section{Discussion}
Our results demonstrate that postprocessing can have a measurable impact on segmentation accuracy, particularly for challenging tumor subregions such as the ET and RC classes. Strategy 3, which combines class-specific small object replacement and targeted filtering, consistently improves the average Dice and HD95 metrics across multiple classes. These improvements suggest that thoughtful refinement of predicted masks can address common sources of segmentation error, such as small false positives. However, despite these gains in average metric values, our analysis using the BraTS ranking system reveals a more nuanced picture. While Strategy 3 generally achieves the best average metrics, these improvements do not translate into better rankings under the BraTS scoring protocol. In fact, Strategy 3 receives the worst average global rank across both cross-validation and validation set evaluations. This underperformance stems from the fact that while Strategy 3 improves a subset of outlier cases, it degrades performance on a larger number of patients. In other words, the strategy harms more predictions than it helps, highlighting the importance of evaluating consistency across the entire cohort rather than focusing solely on aggregate metrics.

By contrast, Strategy 1 is a more straightforward and more targeted postprocessing approach that exclusively addresses small false positives in the RC class. It achieves the best global average rank in our five-fold cross-validation because, unlike Strategy 3, it improves more cases than it degrades. Specifically, for the RC Dice coefficient, Strategy 1 improves 589 cases, leaves 2079 unchanged, and degrades 207 cases. For HD95, it improves 153 cases, leaves 2567 unchanged, and degrades 152 cases. The higher number of improvements focused on a single class, combined with no negative impact on the other five segmentation classes, explains why Strategy 1 achieves the best average rank. However, on the validation set, Strategy 1 slightly underperforms the baseline predictions. 

Given this outcome and the consistent strength of the baseline across all evaluation settings, we elect to submit the baseline model with no postprocessing as our final entry. While postprocessing strategies like those tested here remain valuable in many applications \cite{patel2024training, kim2021model, chen2022end}, their effectiveness depends heavily on the model’s initial performance and the broader evaluation context. In our case, each model was trained for 10,000 epochs using strong architectural priors (i.e., deep supervision, residual blocks), extensive data augmentation, and a conservative learning rate schedule, yielding models that are already highly optimized.

Figure~\ref{fig:loss-curves}, which plots the training and validation loss curves for a single fold, illustrates the high level of optimization that our training scheme achieves. In this plot, we see that the model converges smoothly, with two distinct drops in the loss curves around epochs 1,000 and 6,000, followed by long plateaus. These inflection points likely mark key learning milestones, such as improved delineation of tumor subregions or the ability to distinguish between pre- and post-operative cases. The stability of the loss after these transitions suggests that the model has reached a mature optimization state, leaving limited room for additional gains through postprocessing. This observation further reinforces our decision to submit the baseline model as our final entry.

From a systems perspective, the modularity of the MIST postprocessing pipeline is crucial for facilitating this analysis. Researchers have the flexibility to experiment with a broad array of transformations without the need to retrain models or alter the underlying Python code, which accelerates experimentation and enhances reproducibility. Our strategy configuration files simplify the process of sharing and comparing various refinement approaches, ensuring that evaluations are both transparent and customizable. Collectively, our findings underscore two key considerations for future segmentation workflows: (1) postprocessing remains a valuable tool for enhancing segmentation quality but must be carefully evaluated against robust ranking systems; and (2) modular frameworks like MIST are vital for enabling flexible experimentation, especially in environments where clinical or benchmark requirements are constantly evolving. MIST is open-source (Apache 2.0 license) and is available on \href{https://github.com/mist-medical/MIST}{GitHub} or \href{https://pypi.org/project/mist-medical/}{PyPI}. Please cite \cite{celaya2022pocketnet} and \cite{celaya2024mist} if you use MIST for your own work.

\section{Acknowledgments}
The Department of Defense supports Adrian Celaya through the National Defense Science \& Engineering Graduate Fellowship Program. This research was partially supported by the Tumor Measurement Initiative through the MD Anderson Strategic Research Initiative Development (STRIDE), NSF-2111147, NSF-2111459, and NIH R01CA195524.

\clearpage
\bibliographystyle{splncs04}
\bibliography{sources}

\end{document}